\documentclass[letterpaper, 10 pt, conference]{ieeeconf}  % Comment this line out if you need a4paper

\IEEEoverridecommandlockouts   

\usepackage{graphics} % for pdf, bitmapped graphics files
\usepackage{epsfig} % for postscript graphics files
\usepackage{times} % assumes new font selection scheme installed
\usepackage{multicol}
\usepackage[bookmarks=true]{hyperref}
\usepackage{array}
\usepackage{hyperref}
\usepackage{amsmath}
\usepackage{cleveref}
\usepackage{paralist}
\usepackage{float}
\usepackage{pdfpages}
\let\labelindent\relax
\usepackage{enumitem}
\usepackage{wrapfig}
\usepackage{booktabs} 
\usepackage{multirow}
\usepackage{hyperref}
\usepackage{url}
\usepackage{amssymb}
\usepackage{graphicx}
\usepackage{CJKutf8}
\usepackage{algorithm}
\usepackage[noend]{algorithmic}
\usepackage{colortbl}
\usepackage{threeparttable}
\usepackage{mdframed}
\usepackage{tcolorbox}
\tcbuselibrary{minted}
\usepackage{blindtext}
\usepackage{multicol}
\usepackage{caption, subcaption}
\usepackage[utf8]{inputenc}
\usepackage{tcolorbox}
\usepackage{listings}
\usepackage{xcolor}
\usepackage{tabularx}
\usepackage{cite}
\usepackage{etoolbox}
\definecolor{customblue}{HTML}{ccf2f5}
\usepackage{soul, color, xcolor}

\newcommand{\ours}[0]{Manual2Skill++}

\title{ \LARGE \bf Manual2Skill++: Connector-Aware General Robotic Assembly from Instruction Manuals via Vision–Language Models
\vspace{-0.3mm}
}

\author{
Chenrui Tie$^{* 1}$ \quad
Shengxiang Sun$^{* 2}$ \quad 
Yudi Lin$^{1}$ \quad 
Yanbo Wang$^{3}$ \quad 
Zhongrui Li$^{1}$ \quad 
Zhouhan Zhong$^{1}$ \quad\\
Jinxuan Zhu$^{1}$ \quad
Yiman Pang$^{1}$ \quad
Haonan Chen$^{1}$ \quad
Junting Chen$^{1}$ \quad
Ruihai Wu$^{4}$ \quad
Lin Shao$^{\dagger}$$^{1}$ \quad  \\
\thanks{*~Equal contribution.}%
\thanks{$\dagger$~Corresponding author: Lin Shao (\texttt{linshao@nus.edu.sg}).}%
\thanks{$^{1}$ School of Computing, National University of Singapore, Singapore.}%
\thanks{$^{2}$ University of Toronto, Toronto, Canada.}%
\thanks{$^{3}$ Zhejiang University, Zhejiang, China.}%
\thanks{$^{4}$ Peking University, Beijing, China.}%
}
\begin{document}
\maketitle
\thispagestyle{empty}
\pagestyle{empty}
\renewcommand{\thefootnote}
{\fnsymbol{footnote}}
%%%%%%%%%%%%%%%%%%%%%%%%%%%%%%%%%%%%%%%%%%%%%%%%%%%%%%%%%%%%%%%%%%%%%%%%%%%%%%%%

\begin{abstract}
Assembly hinges on reliably forming connections between parts; yet most robotic approaches plan assembly sequences and part poses while treating connectors as an afterthought.
Connections represent the foundational physical constraints of assembly execution; while task planning sequences operations, the precise establishment of these constraints ultimately determines assembly success.
In this paper, we treat connections as explicit, primary entities in assembly representation, directly encoding connector types, specifications, and locations for every assembly step. 
Drawing inspiration from how humans learn assembly tasks through step-by-step instruction manuals, we present \ours, a vision-language framework that automatically extracts structured connection information from assembly manuals. 
We encode assembly tasks as hierarchical graphs where nodes represent parts and sub-assemblies, and edges explicitly model connection relationships between components.
A large-scale vision-language model parses symbolic diagrams and annotations in manuals to instantiate these graphs, leveraging the rich connection knowledge embedded in human-designed instructions. 
We curate a dataset containing over 20 assembly tasks with diverse connector types to validate our representation extraction approach, and evaluate the complete task understanding-to-execution pipeline across four complex assembly scenarios in simulation, spanning furniture, toys, and manufacturing components with real-world correspondence. More detailed information can be found at \href{https://nus-lins-lab.github.io/Manual2SkillPP/}
{https://nus-lins-lab.github.io/Manual2SkillPP/}
\end{abstract}
%%%%%%%%%%%%%%%%%%%%%%%%%%%%%%%%%%%%%%%%%%%%%%%%%%%%%%%%%%%%%%%%%%%%%%%%%%%%%%%%
\section{Introduction}

% Assembly fundamentally concerns the reliable connection of individual components into a coherent whole entity. 
% Yet the critical factor, establishing physical connections between parts via proper connectors, has not been well-studied in robotics research. Whether in furniture construction, model building, or industrial manufacturing, the selection and application of appropriate connections ($e.g.$, adhesives, mortise-tenon joints, dowels, screws) to secure parts in precise spatial relationships represents the decisive factor between assembly success and failure. Despite this centrality to real-world assembly, prior research has predominantly focused on part-level task planning~\cite{tian2022assemble,tian2024asap,nagpal2024optimal,tie2025manual2skill} or part-level pose estimation~\cite{li2020learning,zhang2024manual,scarpellini2024diffassemble}, often treating connections as a secondary concern. This oversight severely limits the applicability of existing methods to practical assembly scenarios, where connector selection and usage are indispensable.

Assembly fundamentally concerns the reliable connection of individual components into a coherent whole.
Over the past decades, significant progress has been made in robotic execution and spatial reasoning, ranging from learning precise insertion policies~\cite{spector2022insertionnet} to optimizing part-level motions and poses~\cite{tian2022assemble,tian2024asap,li2020learning,scarpellini2024diffassemble}.
However, these methods predominantly focus on the geometry and dynamics of assembly, often treating the underlying connection relationships as predefined or secondary concerns.
In practical scenarios, such as furniture construction or industrial manufacturing, the success of a task depends on more than just spatial alignment; it requires the informed selection and application of specific connectors ($e.g.$, adhesives, mortise-tenon joints, or screws).
There remains a critical gap in interpreting and utilizing structured connection knowledge: identifying the exact type, quantity, and placement of connectors needed for structural integrity.
This gap severely limits the applicability of existing methods to practical assembly scenarios, where connector selection and usage are indispensable.

\begin{figure}[!t]
    \centering
\includegraphics[width=1.0\linewidth, trim=0 0 4cm 0,clip]{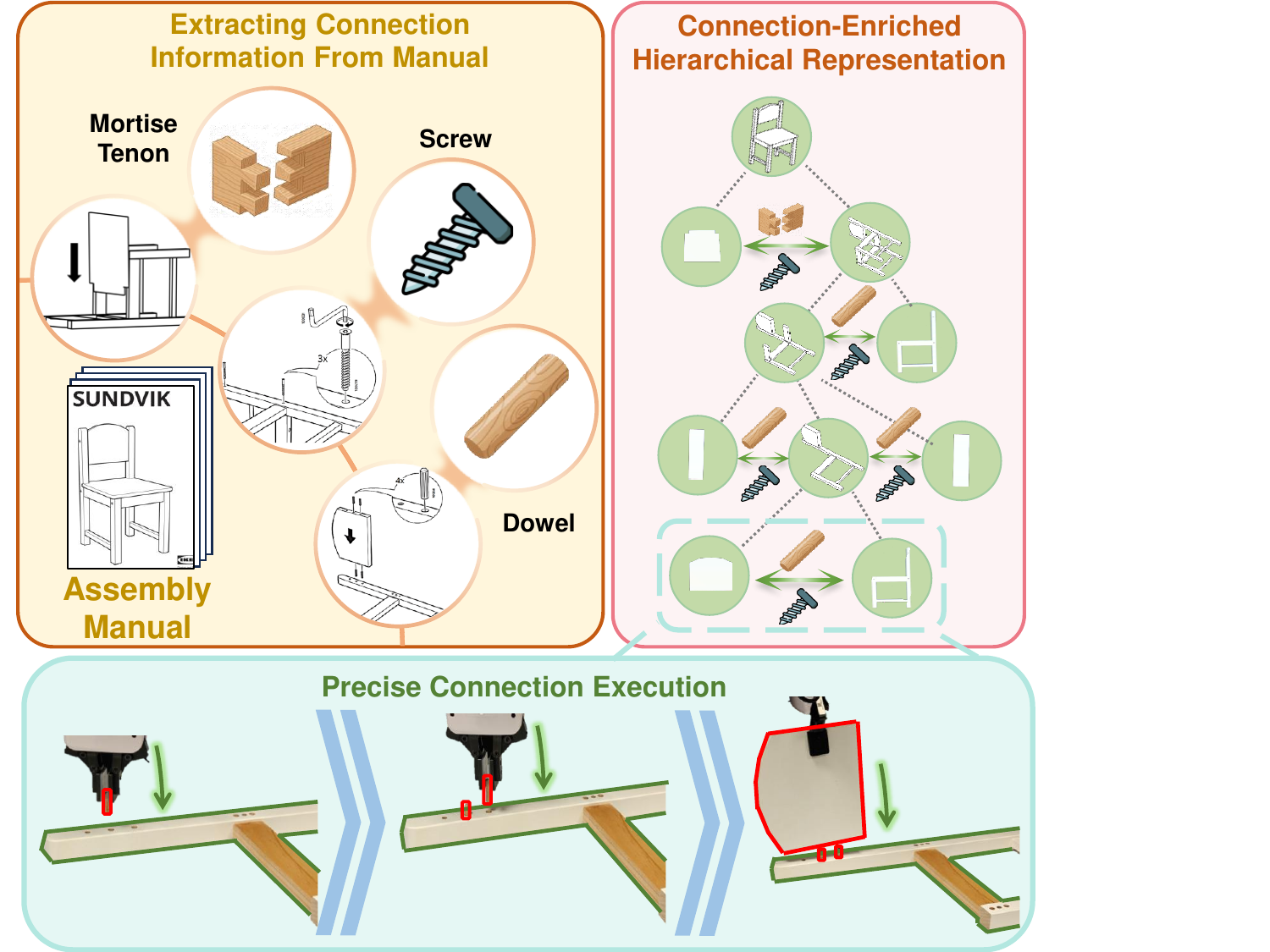}
    \caption{\textbf{Manual2Skill++.} We extract connection information from assembly manuals, build connection-enriched hierarchical graphs encoding part relationships and connector details, then execute precise robotic assembly guided by these connection relations.}
    \label{fig: teaser}
\vspace{-3mm}
\end{figure}

Incorporating connector selection and usage into an assembly framework presents significant challenges that compound the inherent complexity of robotic assembly. 
First, connectors vary widely in form and function; mortise-tenon joints require precise peg insertion, while screws demand controlled rotation and torque, and a single task may involve multiple connection types simultaneously (As shown in~\Cref{fig: teaser}, the assembly of a simple chair involves three types of connection). Second, each connection adds a decision sequence: selecting the connector, locating its placement, and executing the specific insertion or fastening motion. Third, assembly is inherently long-horizon, involving many parts and sub-assemblies, and explicit connector relationships create complex dependencies across the process~\cite{wang2022ikea}. These challenges highlight the need for a structured representation that systematically organizes parts, sub-assemblies, and connectors for comprehensive task understanding and planning.
\clearpage

In everyday life, humans most naturally learn assembly procedures by consulting instruction manuals, whether for IKEA furniture, LEGO models, or hobbyist models.
These manuals encode information of connector selection and placement locations, providing connection knowledge that humans intuitively interpret. 
By parsing sketched illustrations and symbolic annotations, a person can identify not only which parts to assemble, but precisely how they connect through specific connector types and operations. 
Prior work has investigated extraction of task structure and part relationships from manuals~\cite{tie2025manual2skill}, but has largely overlooked this connector-level information, leaving a critical gap in automated assembly understanding.
% Instruction manuals convey rich connection information through abstract symbols and schematic diagrams.
Extracting structured connection information from manuals requires overcoming the ambiguity of sketches and symbolic annotations. To address this, we propose a hierarchical graph representation for assembly tasks. In this graph, leaf and intermediate nodes denote parts and sub-assemblies, while edges explicitly encode connector types, quantities, and spatial constraints. This abstraction bridges high-level task understanding and low-level robotic execution.

Building on this representation, we introduce \ours, a general framework that leverages vision-language models to parse instruction manuals into these hierarchical graphs. By identifying connector correspondences and placement features in a single pipeline, \ours\ enables the direct computation of relative part poses via geometric constraint optimization, achieving significantly higher accuracy than prior learning-based pose estimation methods~\cite{li2020learning,tie2025manual2skill}, thereby providing the precision necessary for robust robotic execution in real-world assembly tasks.

To validate our approach, we curate a dataset of 20+ complex tasks (IKEA furniture, toys, and industrial parts) featuring precise 3D models with annotated connector attachment sites and ground-truth assembly sequences. Furthermore, we develop a simulation benchmark spanning four long-horizon scenarios (As shown in~\Cref{fig: dataset-bench}). Unlike previous works using simplified object models~\cite{heo2023furniturebench,luo2025fmb}, our benchmark incorporates full-scale assets with realistic connection modalities, such as mortise-tenon joints, dowels, and screws, providing a robust testbed for task planning and contact-rich control under realistic assembly constraints with explicit connector complexity.

In summary, we make the following contributions:
\begin{itemize}
\item We propose a graph representation that explicitly models connection relations between assembly components across all assembly tasks, effectively capturing the complexity of assembly tasks while enabling direct computation of part poses through connection constraints.
% , eliminating separate pose estimation pipelines.
\item We introduce \ours, which extracts structured assembly representations from instruction manuals, supported by a curated dataset of over 20 diverse assembly tasks with fine-grained assets and annotations.
\item We develop simulation environments featuring long-horizon assembly tasks with diverse connector types. 
Our benchmark introduces connectors and supports multiple connection operations.
\end{itemize}

\begin{figure*}[!h]
\centering
\includegraphics[width=0.98\linewidth]{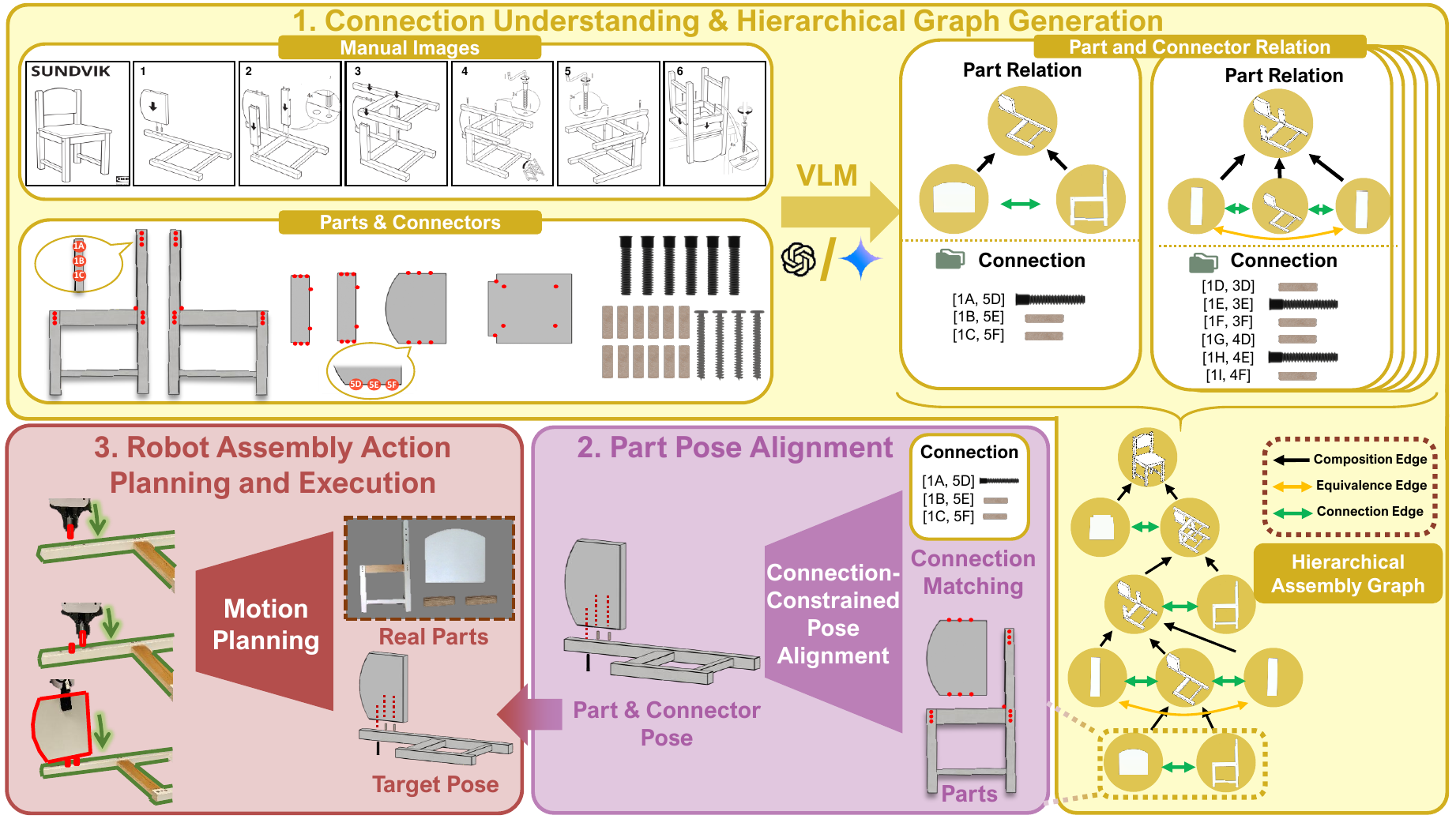} 
\caption{\textbf{Framework Overview.}  
(1) A VLM processes manual images to extract part and connector relations, generating a connection-enriched
hierarchical assembly representation.
(2) The extracted connection constraints guide a geometric optimization process to compute precise target poses for parts and connectors. 
(3) The system executes the assembly by planning and performing robotic actions based on the connection-enriched
hierarchical graph and aligned poses.}
\label{fig: pipeline}
\end{figure*}
\section{related work}
\subsection{Part Assembly}
Part assembly represents a fundamental challenge in robotics, with extensive research exploring how to construct complete objects from individual components~\cite{jones2021automate,lee2021ikea,li2020learning,scarpellini2024diffassemble,wu2023leveraging,tian2024asap,tian2022assemble}. 
Assembly tasks span diverse domains including furniture construction~\cite{lee2021ikea}, toy building~\cite{luo2025fmb} and industrial manufacturing~\cite{jones2021automate}, each presenting unique challenges in terms of part complexity, connection mechanisms, and assembly sequences.
Broadly, we categorize part assembly approaches into \emph{geometric assembly} and \emph{semantic assembly}. 
\emph{Geometric assembly} relies primarily on geometric cues such as surface compatibility, edge features, or shape complementarity to determine part relationships~\cite{wu2023leveraging,sellan2022breaking,du2024generative}. 
These methods excel at puzzle-like tasks where parts fit together based purely on geometric constraints. 
\emph{Semantic assembly} leverages high-level semantic understanding of parts and their functional relationships to guide the assembly process~\cite{jones2021automate,lee2021ikea,li2020learning,tian2022assemble}. This approach is particularly effective for structured objects like furniture, where parts have predefined roles and follow intuitive assembly logic.

Previous research has addressed various aspects of assembly, including motion planning~\cite{suarez2018can}, multi-robot coordination~\cite{knepper2013ikeabot}, pose estimation~\cite{li2020learning,yu2021roboassembly,li2024category}, and sequence planning~\cite{tian2022assemble,tian2024asap}. 
Several datasets and simulation environments have been developed to facilitate research: IKEA furniture datasets~\cite{wang2022ikea} provide 3D models and structured procedures; simulation environments~\cite{lee2021ikea,heo2023furniturebench} enable reproducible evaluation; and specialized benchmarks~\cite{luo2025fmb} offer standardized evaluation protocols.
However, existing approaches predominantly focus on individual subproblems, such as pose estimation or motion planning, while overlooking the critical role of physical connections between parts. Most methods assume simplified part relationships~\cite{luo2025fmb} or rely on oracle mechanics~\cite{lee2021ikea}, neglecting the diverse connection types that determine assembly success in practice. This limits applicability to real-world scenarios where connection selection and execution are paramount.

\subsection{VLM-Guided Robot Learning}
Vision Language Models (VLMs)~\cite{yin2023survey} have been employed in robotics for environment understanding~\cite{huang2024copa}, human–robot interaction~\cite{shi2024yell}, and high-level task planning~\cite{vemprala2024chatgpt}. While end-to-end Vision Language Action models can directly generate robot actions from multimodal inputs~\cite{black2024pi_0,kim2024openvla,team2024octo}, they demand large datasets and often struggle with long-horizon, precise manipulation. An alternative is to use VLMs as reasoning engines, providing structured guidance for task decomposition~\cite{yao2022react}, scene interpretation~\cite{jiang2024roboexp}, and control interfaces~\cite{li2024manipllm,zhao2024large}. Additionally, VLMs have been applied to robot learning tasks such as instruction parsing~\cite{huang2024rekep} and assembly design assistance~\cite{goldberg2024blox}. Building on this paradigm, we introduce \ours, a novel application of VLMs to parse symbolic assembly manuals and extract detailed connection relationships, including types, quantities, and placement constraints, enabling precise, connection-aware robotic assembly operations.

% \subsection{Learning from Demonstrations}
% Learning from demonstration~(LfD) has achieved promising results in acquiring robot manipulation skills~\cite{fu2024mobile,zhu2023viola,chi2023diffusion}. For a broader review of LfD in robotic assembly, we refer to ~\cite{zhu2018robot}.
% The key idea is to learn a policy that imitates the expert's behavior.
% However, previous learning methods often require fine-grained demonstrations, like robot trajectories~\cite{chi2023diffusion} or videos~\cite{kareer2024egomimic,sontakke2024roboclip,jonnavittula2024view}.
% Collecting these demonstrations is often labor-intensive and may not always be feasible.
% Some works propose to learn from coarse-grained demonstrations, like the hand-drawn sketches of desired scenes~\cite{sundaresan2024rt} or rough trajectory sketches~\cite{gu2023rt}.
% These approaches reduce dependence on expert demonstrations and improve the practicality of LfD. However, they are mostly limited to tabletop manipulation tasks and do not generalize well to more complex, long-horizon assembly problems.
% In this work, we aim to extend LfD beyond these constraints by tackling a more challenging assembly task using abstract instruction manuals. 
\section{Problem Formulation}
\label{sec:method-formulation}
Given the 3D models of all involved parts and connectors in an assembly, and its assembly manual, our goal is to generate a physically feasible sequence of robotic assembly actions to aggregate all parts and connectors into a whole object.
We define the manual pages as a set of $N$ images. $\mathcal{I} = \{I_1, I_2, \cdots, I_N\}$, where each image $I_i$ illustrates a specific step in the assembly process, such as the merging of certain parts or sub-assemblies using certain connectors.

The assembly consists of $M$ individual parts $\mathcal{P} = \{P_1, P_2, \cdots, P_M\}$ and $Q$ connectors $\mathcal{C}= \{C_1, C_2, \cdots, C_Q\}$.
A \emph{subassembly} is any partially or fully assembled structure that forms a proper subset of $\mathcal{P} \oplus \mathcal{C}$ ($e.g.$, $\{P_1, P_2, C_1\}$ denotes $P_1, P_2$ are connected via connector $C_1$). A \textit{component} may refer to a part or a subassembly. We define an \textit{attachment point} as the geometric location on a part designated for connector placement, such as protrusions, pin holes, or screw holes (see red points in Figure \ref{fig: pipeline} (1)). Here, we assume access to precise 3D models of all parts, allowing us to know the exact positions of every attachment point. In practice, these high-fidelity models can be obtained via 3D scanning of physical parts or retrieved from manufacturer CAD files~\cite{Koch_2019_CVPR}, making this assumption realistic in real-world assembly settings. Then all the connections can be formulated as binary pairs of attachment points and corresponding connectors.

\section{method}
\label{sec:method}
Our methodology centers on connection modeling as the key to comprehensive assembly understanding. We introduce a hierarchical graph representation that explicitly encodes component connections as a structured task plan (\Cref{sec:method-repr}), and we propose \ours, a VLM framework that automatically constructs these hierarchical graphs from instruction manuals (\Cref{sec:method-generation}). Leveraging the extracted connection constraints, we directly compute relative part poses via optimization, eliminating the need for separate multi‐part pose estimation module (\Cref{sec:method-constraint_matching}). 
This task plan then guides downstream policies for precise insertion, screwing, or snapping operations. To validate the end-to-end pipeline, we develop an Isaac Lab benchmark featuring four complex assembly tasks with diverse connector modalities, demonstrating the necessity and effectiveness of our connector-aware representation (\Cref{sec:method-benchmark}).

\subsection{General Representation for Assembly Tasks}
\label{sec:method-repr}
Our assembly representation employs a hierarchical graph structure $\mathcal{G}$ where leaf nodes correspond to atomic parts, intermediate nodes represent sub-assemblies at various stages, and the root node represents the final assembled product. The graph organizes assembly information across multiple abstraction levels, with each level capturing the granularity appropriate for different reasoning tasks.

We define three types of edges that capture distinct assembly relationships. Composition edges connect parent-child node pairs, indicating that a parent subassembly is formed by aggregating its child components. Equivalence edges link nodes representing identical parts ($e.g.$, the four legs of a table), signifying that these components are interchangeable during assembly. Most importantly, connection edges between sibling nodes represent physical connections that must be established between components through specific connectors, this constitutes the key innovation of our representation. These connection edges carry attributes that encode how the connected nodes are physically joined.
Formally, we have $\mathcal{G} =\{ \mathcal{V}, \mathcal{E}_{\text{comp}}, \mathcal{E}_{\text{eqv}}, \mathcal{E}_{\text{conn}}\}$, where $\mathcal{V}$ denotes the node set, $\mathcal{E}_{\text{comp}}, \mathcal{E}_{\text{eqv}}, \mathcal{E}_{\text{conn}}$ denote three kinds of edge sets.
Each connection edge $e_{\text{conn}}\in\mathcal{E}_{\text{conn}}$
comprises one or more connection instances, since real assemblies often use multiple connectors ($e.g.$, several screws) between the same parts. 
Each connection instance corresponds to an attachment feature pair joined by a connector.
We define the feature of each connection instance as follow:

\begin{table}[!h]
\centering
\caption{Features of Connection Instance}
\label{tab:connection-features}
\begin{tabularx}{\columnwidth}{@{}lX@{}}
\toprule
\textbf{Feature Name} & \textbf{Description} \\
\midrule
Connection Type & Type of connection, such as mortise-tenon joint, dowel, or screw. \\
Connector & Connector identifier (empty for mortise-tenon joints). \\
% Part Relative Pose & 6-DOF relative pose between the two connected nodes. \\
Attachment Feature on Part 1 & Position and normal vector of an attachment point on Part 1’s local frame. \\
Attachment Feature on Part 2 & Position and normal vector of an attachment point on Part 2’s local frame. \\
\bottomrule
\end{tabularx}
\end{table}

An example of the three kinds of edges can be found at~\Cref{fig: pipeline}(1), where the parent-child nodes are linked by composition edge (denoted by black arrow), physically jointed nodes are linked by connection edge (denoted by green arrow), and equivalent nodes are linked by equivalence edge ($e.g$, two identical rods are connected by orange arrow).

\subsection{VLM-Guided Hierarchical Graph Generation}
\label{sec:method-generation}

To initialize the hierarchical graph's nodes, composition edges, and equivalence edges, we employ Manual2Skill's VLM generation stage \cite{tie2025manual2skill}, which uses the manuals to identify parts needed at every assembly step. Manual2Skill++ then introduces the novel framework of generating the connection edges.
% Since each attachment point’s feature (its position and normal vector) is already encoded in the 3D models, constructing the connection edges of the hierarchical graph reduces to identifying the correct pair of attachment points for each connector.
Since we assume access to the 3D model of each part, constructing the connection edges reduces to identifying the correct pairs of attachment points for each connector. This is challenging as it requires correlating low-detail, abstract manual sketches with highly detailed real-world assembly scenes. 
Large parts like panels are straightforward to match, but connectors and attachment points are much smaller and often drawn with minimal detail.
Furthermore, each step typically involves multiple connectors, which must be matched with their corresponding attachment point pairs simultaneously. Determining these pairings is a complex combinatorial problem: from a large set of candidate attachment points, the correct subset needs to be selected and organized into binary pairs, all while handling visual clutter and occlusions in the unstructured manual images.
% Consequently, reasoning about connectors is much more difficult than matching larger structural parts.

We address these challenges by leveraging the reasoning VLM~\cite{comanici2025gemini}. In our pipeline, the VLM iteratively processes each assembly step. For each iteration, the VLM receives two types of visual inputs. The first type is a manual image that encodes the ground-truth attachment point pair for every connector through abstract sketches. The second type is a set of high-fidelity 2D images of the involved components. Rendered from 3D models via Blender, these images present all candidate attachment points and connectors.
% from which the VLM must select the correct pairs based on the manual page.
% For every assembly step, the visual input includes the corresponding manual image, which abstractly presents each connector's attachment points using sketches, and high-fidelity 2D renderings of the involved components with all their attachment points projected from their corresponding 3D models using Blender, which encodes the total potential attachment sites.
The visual inputs are combined with a text input providing domain-specific knowledge about the connector types. From this combined input, the VLM outputs the precise pairs of attachment points to be linked by each connector.

For example, consider the first assembly step in~\Cref{fig: pipeline}(1). The manual input is the first page of the assembly instructions (denoted with step number 1 on the top left). 
The two components involved are the h-shaped side frame (10 red dots representing attachment points) and the curved backrest panel (6 red dots). 
All attachment points are uniquely indexed, though only some are shown in~\Cref{fig: pipeline} for readability. These component inputs show 16 candidate attachment points in total. Given this information, the VLM first infers from the manual that the step requires two wooden dowels. It then reasons over the candidate points and predicts the exact pairs for each dowel: [1B, 5E] and [1C, 5F]. 
% These predicted pairs are depicted in the “Part and Connector Relation” panel of~\Cref{fig: pipeline}(1). 
The VLM repeats this for all subsequent steps to predict attachment point pairings for the other required connectors. These predicted pairings collectively form the connection edges of the hierarchical graph. 
% \crtie{Maybe describe the generation process of the whole graph, other than only the connection edges}
% \lin{I find it difficult to understand the process. It would be more imnformative if you could provide an exampel maybe use the chair shown in Fig 2.}

We implement this input-output process via a two-stage prompting strategy.
In Stage 1, the VLM uses the manual input to estimate the number and type of connectors that are placed on each component. For the chair's first step, the output is \{“h-shaped side frame”: [2, “dowel”], “curved backrest panel”: [2, “dowel”]\}, indicating that two attachment points on the side frame will be paired with two attachment points on the backrest using 2 dowels. Obtaining this high-level information restricts the subsequent search of a connector's possible attachment points to a specific component rather than all components in the assembly, drastically reducing the combinatorial complexity of matching attachment points into binary pairs.
In Stage 2, the VLM identifies precise attachment point pairs for each connector using the output of Stage 1, along with the same manual and component images as input.
% It outputs the final pairings of attachment points for the current step, which directly form a part of the connection edges in the hierarchical graph. 
For example, in the chair's first assembly step, after determining that there should be two dowels connecting the two components, Stage 2 outputs the two exact pairs, [1B, 5E] and [1C, 5F].  
Repeating both stages across all assembly steps yields the complete hierarchical graph.

By extracting the connector information from manuals and formulating the assembly process into a structured hierarchical graph, \ours\ achieves general task understanding for a spectrum of assembly tasks and empowers downstream assembly execution.

\subsection{Part Pose Alignment via Connection Matching}
\label{sec:method-constraint_matching}

% Accurate part pose estimation is critical for successful assembly execution. Previous approaches typically assume ground-truth part poses are available\cite{tian2022assemble,chen2021planning}, which requires extensive additional sensing and calibration infrastructure that is impractical in many scenarios. 
% Alternatively, most recent work formulates this as a multi-part pose regression problem using learning-based methods\cite{li2020learning,scarpellini2024diffassemble,wu2023leveraging,li2024category}. 
% However, these approaches generally suffer from centimeter-level errors that are prohibitive for precise connection operations such as peg insertion or screw fastening.

In our framework, we address part pose estimation through direct connection matching. 
Given that precise attachment points are known and their pairwise correspondences are extracted from assembly manuals, we can directly compute the relative poses between connected components through geometric constraint satisfaction. This approach significantly improves pose alignment accuracy, achieving the precision necessary for reliable assembly operations.
Formally, for each connection edge $e_{\text{conn}}$, the attachment features are denoted as
${[(x^a_1,n^a_1), (x^b_1,n^b_1)],...,[(x^a_k,n^a_k), (x^b_k,n^b_k)]}$
, where $k$ is the number of connection instances ($e.g.$, if two parts are connected by 3 screws, $k=3$) between part $a$ and part $b$, $x\in \mathbb{R}^3$ is the position of attachment, $n\in \mathbb{R}^3$ is the normal unit vector of attachment.
We formulate the pose alignment as the following optimization problem:

\begin{equation}
\begin{aligned}
\min_{R, t} \quad & \sum_{i=1}^{k} \left( | Rx^a_i + t - x^b_i |^2 + \alpha | Rn^a_i + n^b_i |^2 \right)
\end{aligned}
\end{equation}

where $R \in SO(3)$ is the rotation matrix and $t \in \mathbb{R}^3$ is the translation vector that transforms part $a$ to align with part $b$. The first term ensures attachment positions coincide, while the second term enforces that attachment normals are collinear and opposite ($i.e.$, $Rn^a_i = -n^b_i$). The weight $\alpha$ balances position and orientation alignment.
This constraint-based approach achieves millimeter-level accuracy required for reliable connection operations.

\subsection{Robotic Assembly Benchmark in Isaac Lab}
\label{sec:method-benchmark}

% Previous benchmarks built long-horizon assembly tasks but typically rely on simplified custom models~\cite{heo2023furniturebench,luo2025fmb} or oracle mechanics for part aggregation~\cite{lee2021ikea}, often overlooking explicit connector interactions. 
To address the importance of explicit connector-modeling, we present four complex, long-horizon assembly tasks in Isaac Lab, featuring full-scale IKEA furniture and authentic toy models with diverse connection mechanics.

\subsubsection{Task Selection and Complexity} As illustrated in~\Cref{fig: dataset-bench}, our benchmark encompasses the complete assembly of an IKEA chair, shoe shelf, airplane model, and LEGO figure. These tasks span varying complexity levels with different numbers of parts, assembly steps, and connector types, as summarized in~\Cref{tab:bench}. Here the \textit{step} is defined as the number of connection operations, as the joint of two parts may involve multiple connection operations.

\begin{table}[!h]
\centering
\caption{Summary of Four Assembly Tasks}
\label{tab:bench}
\begin{tabular}{lcc|ccc}
\toprule
& & & \multicolumn{3}{c}{\textbf{Connector Types}} \\
\cmidrule(lr){4-6}
\textbf{Task Name} & \textbf{Parts} & \textbf{Steps} & \textbf{Mortise-tenon} & \textbf{Dowels} & \textbf{Screws} \\
\midrule
Shoe Shelf & 4 & 11 & \checkmark  & $\times$ & \checkmark \\
Chair & 6 & 22 & \checkmark & $\checkmark$ & \checkmark \\
LEGO Person & 9 & 8 & \checkmark & $\times$ & $\times$ \\
Plane Model & 11 & 12 & \checkmark & \checkmark & $\times$ \\
\bottomrule
\end{tabular}
\end{table}

\subsubsection{Connection Mechanics Implementation} We design simulation mechanics for three primary connector types. 
\textbf{Mortise-tenon joints} are realized through proximity-based fixed joint creation when relative pose alignment falls within specified thresholds. \textbf{Dowels} similarly trigger fixed joint formation upon correct part-to-dowel pose alignment. Similar to mortise-tenon and dowel connections, \textbf{screws} require precise pose alignment to initiate a connection. However, instead of forming a fixed joint immediately, we employ a multi-stage D6 joint mechanism:  initial proximity creates a constrained joint that permits only rotation along its central axis. Incremental translation along this same axis is then unlocked as rotational thresholds are exceeded, realistically simulating screw tightening without complex friction modeling. 
This connector-aware simulation provides a faithful testbed for evaluating assembly planning and control under realistic connection constraints.
\section{experiment}

\begin{figure*}[!htb]
\centering
\includegraphics[width=1.0\linewidth, page=1]{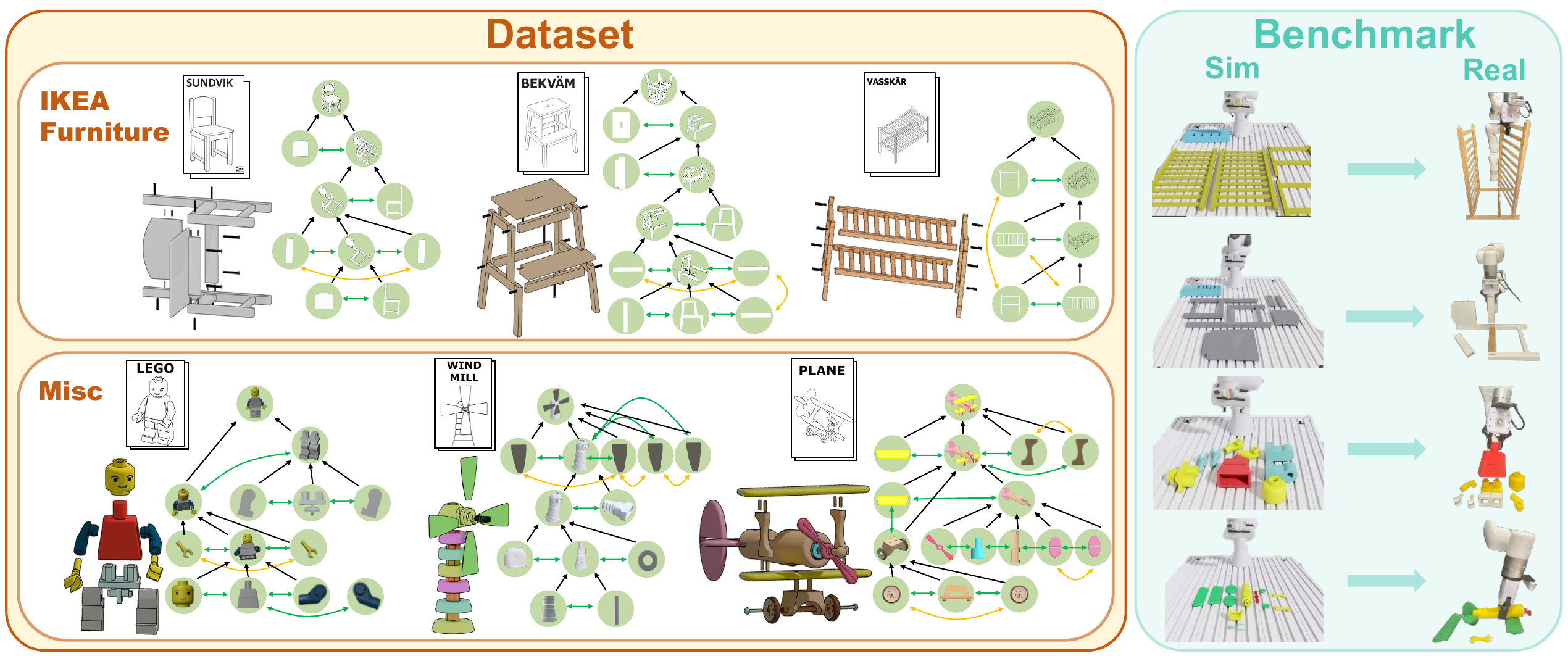} 
\caption{\textbf{Overview of Our Dataset and Benchmark.} The dataset comprises 21 representative assembly tasks, each with high-fidelity 3D part models, manual pages for every step, and fully annotated connection information. From this collection, we selected four tasks and implemented them in Isaac Lab, designing connection mechanics to enable long-horizon assembly with explicit connector operations that directly correspond to real-world procedures.}
\vspace{-2mm}
\label{fig: dataset-bench}
\end{figure*}

In this section, we perform a series of experiments aimed
at addressing the following questions.
\begin{itemize}
    \item Q1: Can our hierarchical graph representation generalize across diverse assembly domains and complexity levels? (\Cref{sec:exp-graph}) 
    \item Q2: How effectively can \ours\ extract the representation from manuals? (\Cref{sec:exp-graph}) 
    \item Q3: Can our connection-matching pose alignment achieve the precision required for practical assembly tasks? (\Cref{sec:exp-pose})
    \item Q4: Are our graph representation and pose alignment sufficient to guide downstream connection execution? (\Cref{sec:exp-bench})
    % \item Q5: Can \ours\ and the associated simulation benchmark be directly applied to and validated on physical, real-world assembly tasks? (\Cref{sec:exp-real})
\end{itemize}

\subsection{Hierarchical Graph Generation}
\label{sec:exp-graph}

% \crtie{left for Shengxiang}
\subsubsection{Dataset}
To validate the universality of our assembly representation and generalization ability of \ours, we curate a dataset covering 21 diverse assembly tasks. It includes 11 IKEA furniture items sampled from the IKEA-Manuals dataset~\cite{wang2022ikea}, and 10 toy models, LEGO sets, and manufacturing components sourced from the Internet. 
For furniture items, we use the original IKEA manual, while for other assembly tasks, we generate instruction manuals using Blender's Freestyle functionality. 

For each task, we manually annotate the 3D positions of all attachment points on every part. To prepare VLM inputs, we projected the 3D models of components and their annotated attachment points into 2D renderings for each assembly step, and we assign unique IDs to each attachment point. The final ground truth for each step consists of the correct pairings of these attachment points for every connector.
% Together, these items cover all 7 types of connector scenarios described in \ref{sec:method-generation} and are diverse in both shape and assembly steps.

Our dataset features a wide range of assembly tasks requiring complex reasoning. On average, a task involves assembling 6 parts over 4 steps. For each step, the model must correctly select and pair around 6 attachment points from a set of 12 possible choices. The dataset also includes high-complexity instances, such as tasks with up to 11 parts, 8 connection steps, or the need to select and pair 12 points from a set of 84 choices.
The fact that every task can be fully described by our connection-enriched hierarchical representation demonstrates its broad applicability and expressiveness. \Cref{fig: dataset-bench} shows 6 examples from our collection.
\subsubsection{Evaluation Metric}
We evaluate the VLM's ability to understand assembly manuals by the accuracy of its predicted attachment point pairs from Stage 2, which are critical for downstream pose estimation and connection execution tasks. Predictions are compared against ground-truth annotations for each assembly step under two complementary settings:
\begin{itemize}
    \item Set Matching (Points Selected): We treat the task as identifying the correct \textit{set of attachment points} that receives connectors. A prediction is correct if it identifies the right points, regardless of how they are paired.
    \item Pair Matching (Connections Made): This setting evaluates the task as identifying the correct \textit{unordered pairs of attachment points} that form individual connections. This is a stricter measure of whether the exact connections are inferred.
\end{itemize}

For each setting, We report the \textit{success rate}, where a single assembly step is considered successful if the predicted set and pairs perfectly match their ground-truth counterparts. We also calculate standard \textit{F1-score} to quantify the overlap between predicted and ground-truth sets or pairs. An attachment point from the predicted set or an attachment point pair from the predicted pairs is a true positive if it matches a ground-truth entry otherwise
it is a false positive; any ground-truth entry not predicted is
a false negative. Parts with equivalent edge are considered as identical.
Final performance is reported as the F1-score and success rate averaged across all 85 assembly steps in the dataset.

% \begin{align*}
% \text{Precision} &= \frac{\text{TP}}{\text{TP} + \text{FP}}, \quad
% \text{Recall} = \frac{\text{TP}}{\text{TP} + \text{FN}}, \
% \text{F1-score} &= 2 \cdot \frac{\text{Precision} \cdot \text{Recall}}{\text{Precision} + \text{Recall}}.
% \end{align*}

\subsubsection{Baseline}
We compare our VLM-based method against two heuristic approaches, and two ablation settings.
\begin{itemize}
    \item Random Sampling: Randomly generates connection pairs from available choices. This evaluates the proportion of trivial assembly steps in our dataset.
    \item Geometric Matching: An OpenCV-based heuristic that identifies attachment points in manuals by referencing manually chosen template images and thresholds. It can only perform Set Matching and completely fails at Pair Matching due to a lack of relational reasoning.
    \item Incomplete Manual: To evaluate how the performance of our method is affected by incomplete or skipped manuals, we randomly replace 1 assembly step with a blank manual image and let VLM infer assembly representation from its prior knowledge and past context.
    \item GPT: We compare the performance of reasoning VLM (Gemini-2.5-Pro ~\cite{comanici2025gemini}), against non-reasoning models that use much fewer tokens (GPT-4o~\cite{hurst2024gpt}). 
    % We use the same two-stage pipeline described in \ref{sec:method-generation} for the two VLM comparisons.
\end{itemize}
\subsubsection{Result}

\begin{table}[ht]
\centering
\begin{threeparttable}
\caption{Result of Assembly Graph Extraction ($\uparrow$)}
\label{tab:full_asssembly_plan}
\renewcommand{\arraystretch}{1.2}
\begin{tabular}{@{}lcc|cc@{}}
\toprule
\textbf{Method} & \multicolumn{2}{c|}{\textbf{Pairs}} & \multicolumn{2}{c}{\textbf{Set}} \\
\cmidrule(lr){2-3} \cmidrule(lr){4-5}
& F1 & Success & F1 & Success \\
\midrule
Geometric Matching & 0.00 & 0.00 & 0.02 & 0.02 \\
Random Sampling & 11.99 & 2.35 & 53.51 & 9.41 \\
Incomplete Manual (GPT) & 36.06 & 22.35 & 75.82 & 41.18 \\
Ours (GPT) & 38.58 & 28.24 & 74.55 & 45.88 \\
Incomplete Manual (Gemini) & 56.71 & 41.18 & 82.68 & 63.53 \\
\rowcolor{customblue}
\textbf{Ours (Gemini)} & \textbf{63.42} & \textbf{49.41} & \textbf{86.06} & \textbf{69.41} \\
\bottomrule
\end{tabular}
\end{threeparttable}
\end{table}

Table \ref{tab:full_asssembly_plan} shows that our method substantially outperforms the random sampling heuristic, with over 51\% and 47\% gains in F1 and success rate for Pair Matching, and over 32\% and 60\% for Set Matching. These results highlight that the assembly steps in our dataset are largely non-trivial and cannot be solved by chance. Similarly, the near-zero success of Geometric Matching highlights that traditional methods fail to generalize across diverse geometries, even with carefully chosen thresholds and templates. \ours\ bridges this gap by leveraging VLM-derived commonsense reasoning to accurately extract complex assembly relationships from abstract manuals. Despite longer runtimes (averaging 2-3 minutes per assembly step) for reasoning VLMs, they exceed non-reasoning VLMs by over 20\% across most metrics, underscoring the dataset's demand for advanced visual-spatial reasoning. Finally, strong performance under the “Incomplete Manual” condition showcases our approach’s robustness: even when a manual step is omitted, both models correctly infer the missing connection information to complete the assembly graph. These findings confirm the viability of our framework to advance automated assembly understanding in a spectrum of assembly tasks.

\subsection{Pose Alignment}
\label{sec:exp-pose}

\subsubsection{Baseline}
We evaluate the performance of our method on our dataset. 
We compare our method with two baselines: SingleImage~\cite{li2020learning} and Manual2Skill~\cite{tie2025manual2skill}.
% \begin{itemize}
%     \item SingleImage~\cite{li2020learning}: A pipeline for single-image guided object pose estimation that regresses 6D poses from point cloud inputs.
%     \item Manual2Skill~\cite{tie2025manual2skill}: A GNN-based framework that estimates multi-part poses using point cloud regression guided by manual images.
% \end{itemize}

Since our method assumes access to precise attachment feature locations, we provide both baselines with masked point clouds where attachment points are highlighted, ensuring fair comparison under equivalent conditions.

\subsubsection{Evaluation Metric}
We adopt comprehensive evaluation metrics to assess the pose alignment, following~\cite{tie2025manual2skill}, we report Geodesic Distance (GD), Root Mean Squared Error (RMSE), Chamfer Distance (CD) and Part Accuracy (PA).
% \begin{itemize}
%     \item Geodesic Distance (GD), which measures the shortest path distance on the unit sphere between the predicted and ground-truth rotations.
%     \item Root Mean Squared Error (RMSE), which measures the Euclidean distance between the predicted and ground-truth poses.
%     \item Chamfer Distance (CD), which calculates the holistic distance between the predicted and the ground-truth point clouds.
%     \item Part Accuracy (PA), which computes the Chamfer Distance between the predicted and the ground truth point clouds; if the distance is smaller than 0.01m, we count this part as ``correctly placed".
% \end{itemize}

\begin{table}[!h]
    \centering
    \setlength\tabcolsep{4pt} % Consistent spacing
    \renewcommand{\arraystretch}{1.2} % Improved row spacing
    \begin{threeparttable}
    \captionsetup{width=\linewidth}
    \caption{Result of Pose Alignment}
    \label{tab:pose}
    \begin{tabular}{@{}lcccccc@{}} % 7 columns total
    \toprule
& GD$\downarrow$ & RMSE$\downarrow$ & CD$\downarrow$ & PA$\uparrow$ \\ 
    \midrule
    SingleImage& 1.86 & 0.2653 & 0.418 & 0.043 & \\
    Manual2skill& 0.76 & 0.0880 & 0.065 & 0.254 & \\
    \rowcolor{customblue} \textbf{Ours} & \textbf{0.03} & \textbf{0.0013} & \textbf{0.005} & \textbf{0.944}  & \textbf{}\\ 
    \bottomrule
    \label{tab:pose}
        \end{tabular}
    \end{threeparttable}
    \vspace{-2mm}
\end{table}

\subsubsection{Result}

The pose alignment results (\Cref{tab:pose}) reveal that the strongest baseline methods still incur centimeter-level errors, which are insufficient for precise connection operations. In contrast, our constraint‐matching approach consistently achieves millimeter-level alignment accuracy, meeting the stringent precision demands of everyday assembly tasks. 
This substantial improvement underscores the effectiveness of our connection-aware representation: by explicitly modeling physical joint constraints, we enable dramatically more accurate part alignment, thereby bridging the gap between high‐level assembly planning and execution.

\subsection{Connection Execution}
\label{sec:exp-bench}

\subsubsection{Experiment Setup}
To evaluate whether our hierarchical graph representation and constraint-based pose alignment suffice for end-to-end assembly, we decompose each of the four benchmark tasks (Chair, Shoe Shelf, Plane, LEGO Figure) into a sequence of connection operations (insertion for mortise-tenon and dowel; screw tightening for screw) by traversing the corresponding hierarchical graph. We enforce a top-down insertion constraint to ground the relative poses into the world frame: each operation fixes one part on the table and initializes the other part or connector 2 cm above its aligned pose along the global Z-axis.

To quantitatively assess each connection operation, we define a trial as successful if the pose error of the held component relative to its ground-truth pose falls below two thresholds: a rotation error of $\varepsilon_R=0.05$ rad and a translation error of $\varepsilon_t=0.2mm$. These thresholds were carefully calibrated to guarantee that any trial meeting both criteria corresponds to a true insertion or tightening event, $i.e.$, the peg is fully inserted into its designated hole rather than merely contacting or bypassing it. By enforcing this strict metric, we ensure that every recorded success reflects a genuinely correct peg insertion or screw-tightening operation.

\subsubsection{Connection Strategy Evaluation}
Under this setup, we compare three connection strategies:

\begin{itemize}
    \item Random Search: The held part is lowered vertically until the distance to the target hole stops decreasing, then constant downward pressure with small lateral perturbations.
    % is applied to overcome local friction and contact barriers near the correct location.
    \item Grid Search:  A 1\,cm side-length grid with 2\,mm resolution is traversed in an S-shape over the target slot. Insertion is attempted at each grid point, with small perturbations to correct misalignment or tilt-induced friction upon jamming.
    \item Force-Position Hybrid: Extends Grid Search by using a tri-axial force sensor to detect contact forces, triggering a fine insertion phase where displacement compensation based on magnitude and direction of lateral force guide targeted corrections, greatly improving success in tight tolerance scenarios.

\end{itemize}

\begin{table}[!h]
    \centering
    \setlength\tabcolsep{4pt} % Consistent spacing
    \renewcommand{\arraystretch}{1.2} % Improved row spacing
    \begin{threeparttable}
    \captionsetup{width=\linewidth}
    \caption{Result of Connection Policy ($\uparrow$)} 
    \label{tab:parts_performance}
    \begin{tabular}{@{}lcccccc@{}} % 7 columns total
    \toprule
    & Chair& Shoe Shelf & Plane& LEGO Figure& \\ 
    \midrule
    Random Search  & 0.039 & 0.033 & 0.015 & 0.157 & \\
    Grid Search & 0.733 & 0.760 & 0.687 & 0.707 & \\
    % PPO &  &  & & & \\
    \rowcolor{customblue} \textbf{Force-Position hybrid} & \textbf{0.767} & \textbf{0.816} & \textbf{0.737} & \textbf{0.750} & \textbf{}\\ 
    \bottomrule
    \label{tab:assembly_plan}
    \end{tabular}
    \end{threeparttable}
\end{table}

\subsubsection{Result} 

The results in~\Cref{tab:parts_performance} represent averages connection success rate across all sub-tasks within each item, each connection operation is repeated 100 times with 3mm uniform initial pose perturbation.
Random Search consistently fails across all tasks. 
Grid Search demonstrates moderate performance with success rates between 68-76\%, but remains sensitive to pose uncertainties and connector tolerances. Force-Position Hybrid strategy achieves the highest reliability, with success rates ranging from 73.7\% to 81.6\% across all tasks.
These results demonstrate two key findings: first, our hierarchical graph representation correctly determines the assembly sequence for complex multi-step tasks. More importantly, our connection-constrained pose alignment achieves millimeter-level precision that enables robust execution through simple local search strategies. 
The consistent performance across diverse connection scenarios validates that our framework provides a reliable foundation for automating complex assembly tasks.

\section{conclusion}

This paper presents \ours, a vision-language framework that explicitly models connector relationships in robotic assembly. By extracting structured hierarchical graphs from manuals, our approach achieves millimeter-level pose alignment, enabling robust execution across complex tasks. We bridge the gap in connection understanding by curating a diverse dataset with explicit annotations and establishing a realistic simulation benchmark with multiple connector modalities. This work provides the community with the tools to develop and evaluate methods on authentic, contact-rich assembly scenarios. Ultimately, \ours\ lays the groundwork for future advances in connector-aware manipulation, multimodal task understanding, and the development of truly generalizable assembly systems.
% \addtolength{\textheight}{-12cm}   

%%%%%%%%%%%%%%%%%%%%%%%%%%%%%%%%%%%%%%%%%%%%%%%%%%%%%%%%%%%%%%%%%%%%%%%%%%%%%%%%
\patchcmd{\thebibliography}{\clubpenalty}{\clubpenalty\footnotesize}{}{}
\bibliographystyle{IEEEtran}
\bibliography{references}

@article{tie2025manual2skill,
  title={Manual2skill: Learning to read manuals and acquire robotic skills for furniture assembly using vision-language models},
  author={Tie, Chenrui and Sun, Shengxiang and Zhu, Jinxuan and Liu, Yiwei and Guo, Jingxiang and Hu, Yue and Chen, Haonan and Chen, Junting and Wu, Ruihai and Shao, Lin},
  journal={arXiv preprint arXiv:2502.10090},
  year={2025}
}

@article{black2024pi_0,
  title   = {pi\_0: A Vision-Language-Action Flow Model for General Robot Control},
  author  = {Black, Kevin and Brown, Noah and Driess, Danny and Esmail, Adnan and Equi, Michael and Finn, Chelsea and Fusai, Niccolo and Groom, Lachy and Hausman, Karol and Ichter, Brian and others},
  journal = {arXiv preprint arXiv:2410.24164},
  year    = {2024}
}

@article{goldberg2024blox,
  title   = {Blox-Net: Generative Design-for-Robot-Assembly Using VLM Supervision, Physics Simulation, and a Robot with Reset},
  author  = {Goldberg, Andrew and Kondap, Kavish and Qiu, Tianshuang and Ma, Zehan and Fu, Letian and Kerr, Justin and Huang, Huang and Chen, Kaiyuan and Fang, Kuan and Goldberg, Ken},
  journal = {arXiv preprint arXiv:2409.17126},
  year    = {2024}
}

@article{heo2023furniturebench,
  title   = {Furniturebench: Reproducible real-world benchmark for long-horizon complex manipulation},
  author  = {Heo, Minho and Lee, Youngwoon and Lee, Doohyun and Lim, Joseph J},
  journal = {arXiv preprint arXiv:2305.12821},
  year    = {2023}
}

@article{huang2024copa,
  title   = {Copa: General robotic manipulation through spatial constraints of parts with foundation models},
  author  = {Huang, Haoxu and Lin, Fanqi and Hu, Yingdong and Wang, Shengjie and Gao, Yang},
  journal = {arXiv preprint arXiv:2403.08248},
  year    = {2024}
}

@article{huang2024rekep,
  title   = {Rekep: Spatio-temporal reasoning of relational keypoint constraints for robotic manipulation},
  author  = {Huang, Wenlong and Wang, Chen and Li, Yunzhu and Zhang, Ruohan and Fei-Fei, Li},
  journal = {arXiv preprint arXiv:2409.01652},
  year    = {2024}
}

@article{jiang2024roboexp,
  title   = {RoboEXP: Action-Conditioned Scene Graph via Interactive Exploration for Robotic Manipulation},
  author  = {Jiang, Hanxiao and Huang, Binghao and Wu, Ruihai and Li, Zhuoran and Garg, Shubham and Nayyeri, Hooshang and Wang, Shenlong and Li, Yunzhu},
  journal = {arXiv preprint arXiv:2402.15487},
  year    = {2024}
}

@article{jones2021automate,
  title     = {Automate: A dataset and learning approach for automatic mating of cad assemblies},
  author    = {Jones, Benjamin and Hildreth, Dalton and Chen, Duowen and Baran, Ilya and Kim, Vladimir G and Schulz, Adriana},
  journal   = {ACM Transactions on Graphics (TOG)},
  volume    = {40},
  number    = {6},
  pages     = {1--18},
  year      = {2021},
  publisher = {ACM New York, NY, USA}
}

@article{kim2024openvla,
  title   = {OpenVLA: An Open-Source Vision-Language-Action Model},
  author  = {Kim, Moo Jin and Pertsch, Karl and Karamcheti, Siddharth and Xiao, Ted and Balakrishna, Ashwin and Nair, Suraj and Rafailov, Rafael and Foster, Ethan and Lam, Grace and Sanketi, Pannag and others},
  journal = {arXiv preprint arXiv:2406.09246},
  year    = {2024}
}

@inproceedings{knepper2013ikeabot,
  title        = {Ikeabot: An autonomous multi-robot coordinated furniture assembly system},
  author       = {Knepper, Ross A and Layton, Todd and Romanishin, John and Rus, Daniela},
  booktitle    = {2013 IEEE International conference on robotics and automation},
  pages        = {855--862},
  year         = {2013},
  organization = {IEEE}
}

@inproceedings{lee2021ikea,
  title        = {IKEA furniture assembly environment for long-horizon complex manipulation tasks},
  author       = {Lee, Youngwoon and Hu, Edward S and Lim, Joseph J},
  booktitle    = {2021 ieee international conference on robotics and automation (icra)},
  pages        = {6343--6349},
  year         = {2021},
  organization = {IEEE}
}

@inproceedings{li2020learning,
  title        = {Learning 3d part assembly from a single image},
  author       = {Li, Yichen and Mo, Kaichun and Shao, Lin and Sung, Minhyuk and Guibas, Leonidas},
  booktitle    = {Computer Vision--ECCV 2020: 16th European Conference, Glasgow, UK, August 23--28, 2020, Proceedings, Part VI 16},
  pages        = {664--682},
  year         = {2020},
  organization = {Springer}
}

@inproceedings{li2024category,
  title     = {Category-level multi-part multi-joint 3d shape assembly},
  author    = {Li, Yichen and Mo, Kaichun and Duan, Yueqi and Wang, He and Zhang, Jiequan and Shao, Lin},
  booktitle = {Proceedings of the IEEE/CVF Conference on Computer Vision and Pattern Recognition},
  pages     = {3281--3291},
  year      = {2024}
}

@inproceedings{scarpellini2024diffassemble,
  title     = {DiffAssemble: A Unified Graph-Diffusion Model for 2D and 3D Reassembly},
  author    = {Scarpellini, Gianluca and Fiorini, Stefano and Giuliari, Francesco and Moreiro, Pietro and Del Bue, Alessio},
  booktitle = {Proceedings of the IEEE/CVF Conference on Computer Vision and Pattern Recognition},
  pages     = {28098--28108},
  year      = {2024}
}

@article{sellan2022breaking,
  title   = {Breaking bad: A dataset for geometric fracture and reassembly},
  author  = {Sell{\'a}n, Silvia and Chen, Yun-Chun and Wu, Ziyi and Garg, Animesh and Jacobson, Alec},
  journal = {Advances in Neural Information Processing Systems},
  volume  = {35},
  pages   = {38885--38898},
  year    = {2022}
}

@article{shi2024yell,
  title   = {Yell at your robot: Improving on-the-fly from language corrections},
  author  = {Shi, Lucy Xiaoyang and Hu, Zheyuan and Zhao, Tony Z and Sharma, Archit and Pertsch, Karl and Luo, Jianlan and Levine, Sergey and Finn, Chelsea},
  journal = {arXiv preprint arXiv:2403.12910},
  year    = {2024}
}

@article{suarez2018can,
  title     = {Can robots assemble an IKEA chair?},
  author    = {Su{\'a}rez-Ruiz, Francisco and Zhou, Xian and Pham, Quang-Cuong},
  journal   = {Science Robotics},
  volume    = {3},
  number    = {17},
  pages     = {eaat6385},
  year      = {2018},
  publisher = {American Association for the Advancement of Science}
}

@article{tian2022assemble,
  title     = {Assemble them all: Physics-based planning for generalizable assembly by disassembly},
  author    = {Tian, Yunsheng and Xu, Jie and Li, Yichen and Luo, Jieliang and Sueda, Shinjiro and Li, Hui and Willis, Karl DD and Matusik, Wojciech},
  journal   = {ACM Transactions on Graphics (TOG)},
  volume    = {41},
  number    = {6},
  pages     = {1--11},
  year      = {2022},
  publisher = {ACM New York, NY, USA}
}

@article{wang2022ikea,
  title   = {Ikea-manual: Seeing shape assembly step by step},
  author  = {Wang, Ruocheng and Zhang, Yunzhi and Mao, Jiayuan and Zhang, Ran and Cheng, Chin-Yi and Wu, Jiajun},
  journal = {Advances in Neural Information Processing Systems},
  volume  = {35},
  pages   = {28428--28440},
  year    = {2022}
}

@inproceedings{wu2023leveraging,
  title     = {Leveraging se (3) equivariance for learning 3d geometric shape assembly},
  author    = {Wu, Ruihai and Tie, Chenrui and Du, Yushi and Zhao, Yan and Dong, Hao},
  booktitle = {Proceedings of the IEEE/CVF International Conference on Computer Vision},
  pages     = {14311--14320},
  year      = {2023}
}

@article{yin2023survey,
  title   = {A survey on multimodal large language models},
  author  = {Yin, Shukang and Fu, Chaoyou and Zhao, Sirui and Li, Ke and Sun, Xing and Xu, Tong and Chen, Enhong},
  journal = {arXiv preprint arXiv:2306.13549},
  year    = {2023}
}

@article{yu2021roboassembly,
  title   = {Roboassembly: Learning generalizable furniture assembly policy in a novel multi-robot contact-rich simulation environment},
  author  = {Yu, Mingxin and Shao, Lin and Chen, Zhehuan and Wu, Tianhao and Fan, Qingnan and Mo, Kaichun and Dong, Hao},
  journal = {arXiv preprint arXiv:2112.10143},
  year    = {2021}
}

@article{team2024octo,
  title={Octo: An open-source generalist robot policy},
  author={Team, Octo Model and Ghosh, Dibya and Walke, Homer and Pertsch, Karl and Black, Kevin and Mees, Oier and Dasari, Sudeep and Hejna, Joey and Kreiman, Tobias and Xu, Charles and others},
  journal={arXiv preprint arXiv:2405.12213},
  year={2024}
}

@article{vemprala2024chatgpt,
  title={Chatgpt for robotics: Design principles and model abilities},
  author={Vemprala, Sai H and Bonatti, Rogerio and Bucker, Arthur and Kapoor, Ashish},
  journal={IEEE Access},
  year={2024},
  publisher={IEEE}
}

@article{yao2022react,
  title={React: Synergizing reasoning and acting in language models},
  author={Yao, Shunyu and Zhao, Jeffrey and Yu, Dian and Du, Nan and Shafran, Izhak and Narasimhan, Karthik and Cao, Yuan},
  journal={arXiv preprint arXiv:2210.03629},
  year={2022}
}

@article{zhao2024large,
  title={Large language models as commonsense knowledge for large-scale task planning},
  author={Zhao, Zirui and Lee, Wee Sun and Hsu, David},
  journal={Advances in Neural Information Processing Systems},
  volume={36},
  year={2024}
}

@inproceedings{li2024manipllm,
  title={Manipllm: Embodied multimodal large language model for object-centric robotic manipulation},
  author={Li, Xiaoqi and Zhang, Mingxu and Geng, Yiran and Geng, Haoran and Long, Yuxing and Shen, Yan and Zhang, Renrui and Liu, Jiaming and Dong, Hao},
  booktitle={Proceedings of the IEEE/CVF Conference on Computer Vision and Pattern Recognition},
  pages={18061--18070},
  year={2024}
}

@inproceedings{du2024generative,
  title={Generative 3D Part Assembly via Part-Whole-Hierarchy Message Passing},
  author={Du, Bi'an and Gao, Xiang and Hu, Wei and Liao, Renjie},
  booktitle={Proceedings of the IEEE/CVF Conference on Computer Vision and Pattern Recognition},
  pages={20850--20859},
  year={2024}
}

@inproceedings{tian2024asap,
  title={Asap: Automated sequence planning for complex robotic assembly with physical feasibility},
  author={Tian, Yunsheng and Willis, Karl DD and Al Omari, Bassel and Luo, Jieliang and Ma, Pingchuan and Li, Yichen and Javid, Farhad and Gu, Edward and Jacob, Joshua and Sueda, Shinjiro and others},
  booktitle={2024 IEEE International Conference on Robotics and Automation (ICRA)},
  pages={4380--4386},
  year={2024},
  organization={IEEE}
}

@article{luo2025fmb,
  title={Fmb: a functional manipulation benchmark for generalizable robotic learning},
  author={Luo, Jianlan and Xu, Charles and Liu, Fangchen and Tan, Liam and Lin, Zipeng and Wu, Jeffrey and Abbeel, Pieter and Levine, Sergey},
  journal={The International Journal of Robotics Research},
  volume={44},
  number={4},
  pages={592--606},
  year={2025},
  publisher={SAGE Publications Sage UK: London, England}
}

@InProceedings{Koch_2019_CVPR,
author = {Koch, Sebastian and Matveev, Albert and Jiang, Zhongshi and Williams, Francis and Artemov, Alexey and Burnaev, Evgeny and Alexa, Marc and Zorin, Denis and Panozzo, Daniele},
title = {ABC: A Big CAD Model Dataset For Geometric Deep Learning},
booktitle = {The IEEE Conference on Computer Vision and Pattern Recognition (CVPR)},
month = {June},
year = {2019}
}

@article{hurst2024gpt,
  title={Gpt-4o system card},
  author={Hurst, Aaron and Lerer, Adam and Goucher, Adam P and Perelman, Adam and Ramesh, Aditya and Clark, Aidan and Ostrow, AJ and Welihinda, Akila and Hayes, Alan and Radford, Alec and others},
  journal={arXiv preprint arXiv:2410.21276},
  year={2024}
}

@article{comanici2025gemini,
  title={Gemini 2.5: Pushing the frontier with advanced reasoning, multimodality, long context, and next generation agentic capabilities},
  author={Comanici, Gheorghe and Bieber, Eric and Schaekermann, Mike and Pasupat, Ice and Sachdeva, Noveen and Dhillon, Inderjit and Blistein, Marcel and Ram, Ori and Zhang, Dan and Rosen, Evan and others},
  journal={arXiv preprint arXiv:2507.06261},
  year={2025}
}

@inproceedings{spector2022insertionnet,
  title={Insertionnet 2.0: Minimal contact multi-step insertion using multimodal multiview sensory input},
  author={Spector, Oren and Tchuiev, Vladimir and Di Castro, Dotan},
  booktitle={2022 International Conference on Robotics and Automation (ICRA)},
  pages={6330--6336},
  year={2022},
  organization={IEEE}
}

%%%%%%%%%%%%%%%%%%%%%%%%%%%%%%%%%%%%%%%%%%%%%%%%%%%%%%%%%%%%%%%%%%%%%%%%%%%%%%%%

%%%%%%%%%%%%%%%%%%%%%%%%%%%%%%%%%%%%%%%%%%%%%%%%%%%%%%%%%%%%%%%%%%%%%%%%%%%%%%%%

%%%%%%%%%%%%%%%%%%%%%%%%%%%%%%%%%%%%%%%%%%%%%%%%%%%%%%%%%%%%%%%%%%%%%%%%%%%%%%%%

\end{document}